%% file: main.tex
\documentclass[10pt,twocolumn,letterpaper]{article}

\usepackage{cvpr}              

\usepackage{graphicx}
\usepackage{amsmath}
\usepackage{amssymb}
\usepackage{booktabs}

\usepackage{bbding}

\usepackage[pagebackref,breaklinks,colorlinks]{hyperref}

\usepackage[capitalize]{cleveref}
\crefname{section}{Sec.}{Secs.}
\Crefname{section}{Section}{Sections}
\Crefname{table}{Table}{Tables}
\crefname{table}{Tab.}{Tabs.}

\def\cvprPaperID{ } 
\def\confName{CVPR}
\def\confYear{2024}

\makeatletter
\newcommand{\printfnsymbol}[1]{%
	\textsuperscript{\@fnsymbol{#1}}%
}
\makeatother

\begin{document}

\title{Lazy Visual Localization via Motion Averaging}

\author{
Siyan Dong$^{1,2}$\thanks{Joint first authors} \quad Shaohui Liu$^{2}$\printfnsymbol{1} \quad Hengkai Guo$^{3}$ \quad Baoquan Chen$^{1}$ \quad Marc Pollefeys$^{2}$ \quad  
\\
$^1${Peking University} \quad $^2${ETH Zurich} \quad $^3${ByteDance Inc.} 
}
\maketitle

\begin{abstract}
\input{tex/0_abstract}

\end{abstract}

\input{tex/1_intro}
\input{tex/2_related}

\input{tex/3_method}

\input{tex/4_experiment}

\input{tex/5_conclusion}

\clearpage
{\small
\bibliographystyle{ieee_fullname}
\bibliography{egbib}
}

\end{document}

%% file: tex/0_abstract.tex
Visual (re)localization is critical for various applications in computer vision and robotics. Its goal is to estimate the 6 degrees of freedom (DoF) camera pose for each query image, based on a set of posed database images. Currently, all leading solutions are structure-based that either explicitly construct 3D metric maps from the database with structure-from-motion, or implicitly encode the 3D information with scene coordinate regression models. On the contrary, visual localization without reconstructing the scene in 3D offers clear benefits. It makes deployment more convenient by reducing database pre-processing time, releasing storage requirements, and remaining unaffected by imperfect reconstruction, etc. 
In this technical report, we demonstrate that it is possible to achieve high localization accuracy without reconstructing the scene from the database. The key to achieving this owes to a tailored motion averaging over database-query pairs. 
Experiments show that our visual localization proposal, LazyLoc, achieves comparable performance against state-of-the-art structure-based methods. 
Furthermore, we showcase the versatility of LazyLoc, which can be easily extended to handle complex configurations such as multi-query co-localization and camera rigs.

%% file: tex/1_intro.tex
\section{Introduction}

\input{fig/teaser}

Visual localization is an important step to a large variety of applications, such as augmented and mixed reality~\cite{lynen2015get, ungureanu2020hololens}, robot navigation~\cite{chaplot2020learning, fang2020towards}, autonomous driving~\cite{Geiger2012CVPR, Geiger2013IJRR}, etc. The goal of the task is to estimate the 6 degree-of-freedom (DoF) camera pose for each query image based on a set of calibrated and posed images as database.

The last decade witnesses great success of structure-based visual localization methods. A typical solution is to first explicitly build 3D metric maps with Structure-from-Motion (SfM) \cite{snavely2006photo, wu2013towards, schonberger2016structure}, and then apply keypoint matching \cite{lowe1999object, detone2018superpoint, sarlin2020superglue} to get 2D-3D correspondences, from which the Perspective-n-Point (PnP) solver~\cite{gao2003complete, lepetit2009epnp} can be employed along with a locally optimized RANSAC \cite{fischler1981random,chum2003locally,lebeda2012fixing} to get the final pose estimation. With the rapid development of deep learning, recent methods \cite{shotton2013scene, li2020hierarchical, dong2021robust, brachmann2021visual} propose to employ implicit representation for scene coordinate regression. Specifically, they train neural networks on each specific scene with per-pixel 3D coordinates as the supervision, implicitly encoding the 3D information.

While both classes of methods have achieved promising results on multiple datasets \cite{shotton2013scene, kendall2015posenet} under challenging conditions \cite{Sattler2012BMVC, Sattler2018CVPR, wald2020beyond}, they all require a time-consuming pre-processing step to build 3D metric maps, either explicit with SfM reconstruction or implicit with neural network training. Recent works on implicit representation attempt to either reduce the training time~\cite{cavallari2017fly, cavallari2019real, cavallari2019let, dong2022visual} or make the learned network scene agnostic~\cite{yang2019sanet, tang2021learning}. However, they are hard to generalize to large-scale outdoor scenes.

In this work, we aim to study if one can perform visual localization without pre-built 3D metric maps. For instance, one can obtain database poses from either motion tracking systems \cite{sturm12iros} or laser scans \cite{mastin2009automatic, knapitsch2017tanks}, without acquiring SfM model. Also, one can resort to geo-tagged image collections~\cite{sattler2017large, jafarzadeh2021crowddriven} using GPS and human annotation. Furthermore, recent advances on relative pose estimation \cite{sattler2017large, jafarzadeh2021crowddriven} are agnostic to 3D metric maps and can benefit visual localization with additional constraints. Thus, the pre-built 3D metric maps are not always necessary for acquiring reliable poses of the database images. Moreover, there are several advantages of visual localization without 3D metric maps, which we summarize as follows:

\begin{itemize}\itemsep0pt
    \item \textbf{Reducing database pre-processing time}, since no SfM or neural network training is required.
    \item \textbf{Releasing storage requirements} for 3D metric maps such that no compression \cite{yang2022scenesqueezer} is needed.
    \item \textbf{Not affected by imperfect reconstruction}, as no triangulation is conducted among database images.
    \item \textbf{Convenience for database update} such as insertion, deletion, etc, since no 3D metric map is maintained. 
    \item \textbf{Flexibility} to new descriptors, matchers, and two-view solvers. It is also motivated in MeshLoc~\cite{panek2022meshloc}. 
    \item \textbf{Preservation of privacy} \cite{geppert2020privacy,geppert2022privacy}, since there is no exposed information of 3D reconstruction and the images can be stored as descriptors in distributed solvers.
\end{itemize}

Most related to us, Sattler et al.~\cite{sattler2017large} propose to build local SfM models on the fly with retrieved images. The method is very accurate but extremely inefficient. In recent work~\cite{zhou2020learn}, a method without 3D metric maps is introduced that only runs RANSAC over essential matrix estimation. However, its performance is still far from the structure-based methods. 
In this technical report, we show that one can reach comparable localization performance with structure-based methods without resorting to any pre-built 3D metric maps. 
To achieve this, we adapt the conventional motion averaging problem (Figure~\ref{figure:teaser}) with known database poses as constraints. Also, we build tracks directly from the matches over database-query pairs for pose optimization. 
The resulting visual localization solution, named LazyLoc, achieves comparable accuracy with state-of-the-art methods, and is easier to deploy. 
Furthermore, we show that LazyLoc can be extended to handle complex configurations such as multi-query co-localization and camera rigs. 

%% file: fig/teaser.tex
\begin{figure}[tb]
\centering
	\includegraphics[width=0.99\linewidth]{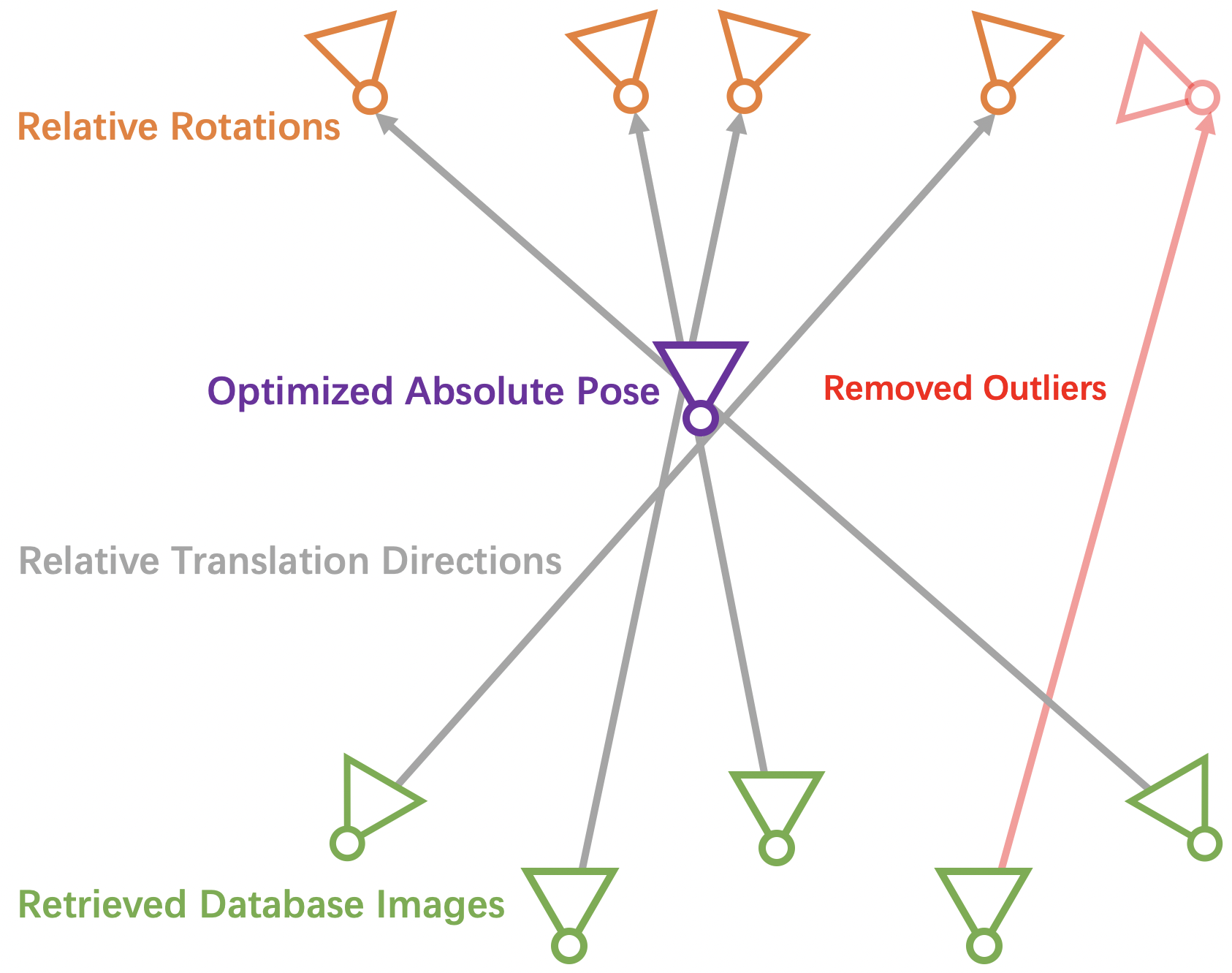}
	\caption{
    Illustration of the proposed visual localization approach via motion averaging. 
    Given the query images, we retrieve database images and apply keypoint matching to compute a set of relative poses. On top of the relative poses, along with fixed known database poses, the absolute pose is optimized using robust rotation averaging, translation averaging, and post optimization. }
	\label{figure:teaser}
\end{figure}

%% file: tex/2_related.tex
\section{Related Work}

\noindent
\textbf{Structure-based visual localization.}
Currently, structure-based approaches are leading in terms of accuracy on most localization benchmarks. Generally, these methods require pre-processing of the database to reconstruct a 3D metric map. During the inference stage, these methods first build correspondences between 2D image pixels and 3D scene points, and then estimate the camera pose using PnP~\cite{gao2003complete, lepetit2009epnp} with RANSAC~\cite{fischler1981random, chum2003locally}. 

To obtain the 2D-3D correspondences, a classical series of works~\cite{sattler2011fast, sattler2012improving, sattler2016efficient} explicitly build SfM models and then perform feature matching between query images and 3D points. Incorporating advanced learning-based image retrieval~\cite{arandjelovic2016netvlad, gordo2017end, revaud2019learning, ge2020self}, keypoint matching~\cite{arandjelovic2012three,detone2018superpoint,dusmanu2019d2,revaud2019r2d2,sarlin2020superglue,sun2021loftr}, and featuremetric alignment~\cite{sarlin2021back}, recent works~\cite{sarlin2019coarse, sarlin2020superglue, sarlin2021back} significantly boost the localization accuracy. Since the learned features are generic enough, most of the aforementioned methods generalize well to both small/large scale and indoor/outdoor scenes. However, the main limitation is the requirement of 3D maps, which take time to be built on top of the databases before inference. Recent work MeshLoc~\cite{panek2022meshloc} makes use of pre-built dense meshes as their 3D maps, which are beyond our focus in this technical report. 

Another popular series is scene coordinate regression, which in general requires re-training each time when comes to a new scene. 
With the help of depth images, the series of works~\cite{shotton2013scene, valentin2015exploiting, cavallari2017fly, cavallari2019real, cavallari2019let} achieve more accurate camera poses than SfM-based methods. 
However, depth images are scarce in real-world scenarios, especially for large-scale outdoor environments. 
Scene coordinate regression with neural networks~\cite{brachmann2017dsac, brachmann2018learning, zhou2020kfnet, li2020hierarchical, brachmann2021visual, huang2021vs} enables queries without depth, and the training time can be significantly reduced~\cite{dong2022visual}. 
There are also impressive methods that make the networks scene agnostic and require no re-training~\cite{yang2019sanet, tang2021learning}. However, the aforementioned methods still need depth images as databases and can not handle large-scale scenes, otherwise not accurate enough.

~\\
\noindent
\textbf{Visual localization without pre-built 3D metric maps.} 
The absolute pose regression (APR) approaches~\cite{kendall2015posenet,kendall2016modelling, kendall2017geometric, sattler2019understanding} make use of end-to-end learning scheme to directly regress camera poses from RGB images. However, as discussed in recent work~\cite{sattler2019understanding}, APR is inherently connected to relative pose regression (RPR)~\cite{laskar2017camera, melekhov2017relative}, and their localization accuracy still lags behind the structure-based methods. 

Besides the pose regression based approaches, there are methods explicitly using multi-view geometry~\cite{sattler2017large, zhou2020learn, bhayani2021calibrated} so that no pre-built 3D metric map is required. 
Specifically, global 3D maps are not always necessary. Instead, local 3D maps can be built on the fly using the images retrieved from the database~\cite{sattler2017large}. 
Zhou et al.~\cite{zhou2020learn} present a method using RANSAC on top of relative poses recovered from essential matrixes. 
Introducing advanced solvers~\cite{bhayani2021calibrated} can significantly reduce camera pose error. 
However, the methods mentioned above are either inefficient or not as accurate as structure-based methods.
In this technical report, we show that by using carefully designed robust motion averaging, one can achieve on-par accuracy with state-of-the-art methods. 

Recently, map-free localization~\cite{arnold2022map} has been studied in a different setup where databases are defined by single images. However, our work introduces a ``counterpart'' under the most widely employed conventional setup. 

\input{fig/pipeline}

\noindent
\textbf{Motion averaging.} 
Motion averaging~\cite{govindu2001combining, hartley2013rotation, chatterjee2013efficient, wilson2014robust, ozyesil2015robust} is widely used in global SfM~\cite{cui2015global, zhu2018very}. Different from incremental SfM~\cite{wu2013towards, schonberger2016structure} alternating between reconstructing 3D metric maps and solving camera poses using 2D-3D correspondences, global SfM solves all the camera poses together based on relative poses before 3D reconstruction, which shares similar spirits to the topic this report discusses, i.e. visual localization without pre-built 3D metric maps. 
Specifically, motion averaging generally involves global rotation averaging~\cite{hartley2003multiple, martinec2007robust, hartley2013rotation, chatterjee2013efficient, arrigoni2014robust, eriksson2018rotation} and translation averaging~\cite{wilson2014robust, ozyesil2015robust, cui2015global, zhuang2018baseline}. In general, solving global translations is more challenging than rotations due to scale ambiguity, and the translation directions depend on relative rotations. In practice, rotation averaging is applied as the first step of motion averaging, which is generally to be robust to outliers \cite{chatterjee2013efficient}. The common practice is to optimize the relative translations with the updated averaged rotations before translation averaging \cite{ozyesil2015robust}. Finally, translation averaging solves for the global translations using the acquired rotations. Different from global SfM, for visual localization, since the poses of database images are known, there is no global scale ambiguity. In this technical report, we reformulate the conventional optimization problem in both stages to support fixed known poses, and introduce a post optimization scheme to further improve the accuracy.

%% file: fig/pipeline.tex
\begin{figure*}[!htb]
\centering
	\includegraphics[width=0.99\linewidth]{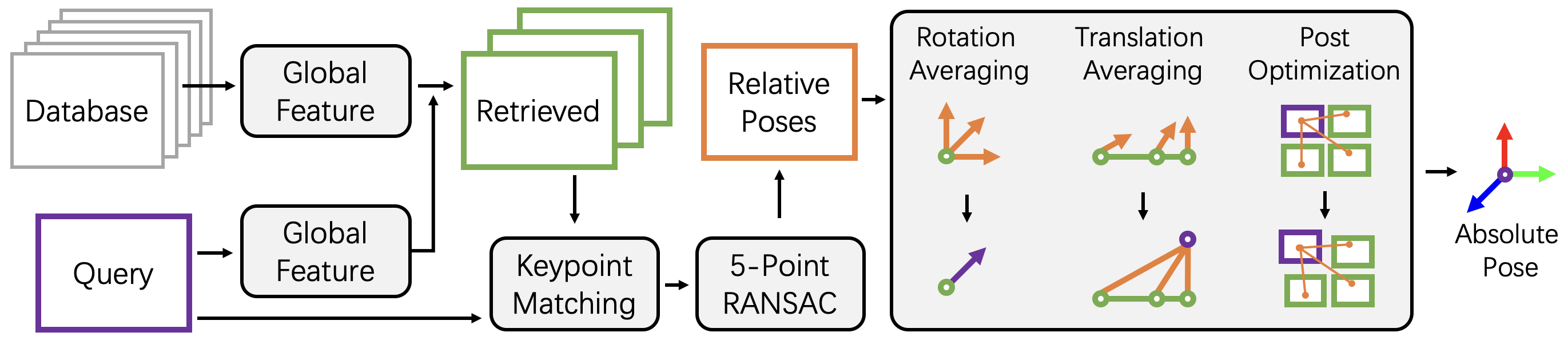}
	\caption{Illustration of the proposed visual localization pipeline LazyLoc. Given a query image, we first extract a whole-image global feature to retrieve the top-$K$ database images which are most similar in appearance. Then, we extract local features and match keypoints in order to solve the relative pose for each database-query pair. Finally, the relative poses are input to a robust motion averaging framework, consisting of rotation averaging, translation averaging, and post optimization, to estimate an absolute pose of the input query image. }
	\label{figure:pipeline}
\end{figure*}

%% file: tex/3_method.tex
\section{Method} 
Given a set of database images with known poses, we aim to localize the query image without resorting to pre-built 3D metric maps. The proposed visual localization pipeline, LazyLoc, is illustrated in Figure \ref{figure:pipeline}. To infer the camera pose of the query image, we first retrieve the top-$K$ similar images from the database based on the global visual appearance. Then, we estimate the relative pose between the query image and each retrieved database image (Section \ref{sec:relpose}). With all the relative poses as input, the motion averaging module consists of rotation averaging (Section \ref{sec:ra}), translation averaging (Section \ref{sec:ta}), and post optimization (Section \ref{sec:ba}). The final output is the 6 DoF absolute pose that is robust to the noisy relative poses.

\subsection{Estimating Two-View Geometry} \label{sec:relpose}
We first discuss here the background and notation of the conventional two-view geometry estimation that is employed in the proposed localization system. 
We denote here the projection matrix of each image $\mathbf{I}_i$ as $\mathbf{P}_i = (\mathbf{R}_i, \mathbf{t}_i)$, with $\mathbf{R}_i \in SO(3)$, $\mathbf{t}_i \in \mathbb{R}^3$ being the rotation and translation respectively. The relative pose between cameras $\mathbf{P}_1$ and $\mathbf{P}_2$ is obtained as:

\begin{equation}
\begin{aligned}
    \mathbf{R}&=\mathbf{R}_2\mathbf{R}_1^\mathrm{T} \\
    \mathbf{t}&= \mathbf{t}_2 - \mathbf{R}_2\mathbf{R}_1^\mathrm{T}\mathbf{t}_1.
\end{aligned}
\label{eq:relative_pose}
\end{equation}

Based on epipolar geometry, we get one constraint from each point correspondence $(\mathbf{p}, \mathbf{p}') \in \mathbb{R}^3 \times \mathbb{R}^3$ on the relative pose: $\mathbf{p'}^\mathrm{T}[\mathbf{t}]_\times \mathbf{R}\mathbf{p}=0$. Since we have access to the camera calibration to both query and database images, we can estimate the relative pose with 5-point algorithm \cite{stewenius2006recent} with robust estimation \cite{fischler1981random}. We use the robust two-view geometry estimator from COLMAP \cite{schonberger2016structure} to acquire the relative pose for each image pair, which consists of a query image and a retrieved database image. We also store all the matches and the inlier indexes for each image pair, which is employed in the post optimization stage.

\subsection{Rotation Averaging} \label{sec:ra}

Given a query image and the recovered relative poses between itself and the retrieved database images, we aim to first solve its rotation by averaging the information from all the recovered relative rotations. Denote $\mathbf{R}_q$ as the absolute rotation of the query image,  $\mathbf{R}_d^i$ as the absolution rotation of the $i$th retrieved database image, and $\mathbf{R}_{qd}^i$ as the recovered relative rotation between the query image and the $i$th retrieved database image. We have the following equation:

\begin{equation}
    \mathbf{R}_{qd}^i = \mathbf{R}_d^i\mathbf{R}_q^\mathrm{T}.
\end{equation}

Since the absolute rotation $\mathbf{R}_d^i$ of the database image is known, we can get one query rotation proposal from each database image by:

\begin{equation}
    \mathbf{R}_q = {\mathbf{R}_{qd}^i}^\mathrm{T}\mathbf{R}_d^i
\end{equation}

So from $k$ valid relative poses from the top-$K$ retrieved images we will get $k$ proposals for the absolute rotation of the query image. A straightforward solution of aggregating the rotation is employed in \cite{zhou2020learn}, where the quaternions are averaged and normalized to get the averaged rotation. However, this strategy can only output an approximated solution and is sensitive to outliers (in \cite{zhou2020learn} the strategy is applied only in local optimization over inlier image pairs).

Inspired by L1-IRLS rotation averaging described in \cite{chatterjee2013efficient}, we here employ a similar strategy by optimizing the rotation in the Lie algebra \cite{govindu2004lie}. Specifically, for each update step, we compute the discrepancy for each image pair as $\Delta R_{qd}^i = {\mathbf{R}_d^i}^\mathrm{T}R_{qd}^iR_q$, map it to the Lie algebra space with $\Delta \omega_{qd}^i=log(\Delta R_{qd}^i)$, then solve for the Lie algebraic update in a linear form $A \Delta \omega_{q}=\Delta \omega_{rel}$ and map it back to the rotation group with the exponential mapping. 
Compared to the original rotation averaging in \cite{chatterjee2013efficient}, in our case here for each step we only solve for the rotation of the query in the Lie algebra, while the rotations of the database images are fixed and thus will not contribute to the matrix $A$ and not included in the Lie algebra update. 
Given $k$ valid relative poses, we first initialize the query rotation using the relative pose with the highest number of inlier matches. Then, we optimize rotation by performing the update in the Lie algebra and mapping back with the L1 norm to solve $A \Delta \omega_{q}=\Delta \omega_{rel}$, which improves robustness to outlier relative poses. Finally, we initialize the solution from the L1 norm update and perform iterative reweighted least squares instead, for the update in the Lie algebra, which outputs a final query rotation.

\subsection{Translation Averaging} \label{sec:ta}

After acquiring the absolute rotation for the query image, we aim to solve its translation, which is an equivalent problem to solving for its camera center. Inspired by the widely used global structure-from-motion system \cite{theia-manual}, we first optimize the relative translation vector for each pair using the solved rotation. We employ the usual strategy to optimize with respect to the epipolar constraints \cite{ozyesil2015robust}. 

With some simple maths from Eq. (\ref{eq:relative_pose}) we get:

\begin{equation}
    \mathbf{R}_2^\mathrm{T}t = \mathbf{R}_2^\mathrm{T}t_2 - \mathbf{R}_1^\mathrm{T}t_1 = \mathbf{C}_1 - \mathbf{C}_2,
\end{equation}

where $\mathbf{C}_1=-\mathbf{R}_1^\mathrm{T}\mathbf{t}_1$ and $\mathbf{C}_2=-\mathbf{R}_2^\mathrm{T}\mathbf{t}_2$ are the camera centers for image 1 and image 2 respectively. 
In our case with the poses of the database images known, we easily get constraints on the camera center $C_q$ of the query image based on the camera center $C_d^i$ for the ith retrieved database image and the relative translation $t_{qd}^i$. As the recovered relative translation from 5-point algorithm \cite{stewenius2006recent} is up-to-scale, the constraint is as follows:

\begin{equation}
    \lambda_i\mathbf{R}_d^\mathrm{T}\mathbf{t}_{qd}^i=\mathbf{C}_q - \mathbf{C}_d
\end{equation}

Each database image contributes two constraints on the global translation of the query image. While translation averaging is generally a hard problem that needs to prevent the scale to collapse to zero, in our case having a set of fixed database images makes it much more stable. In particular, to make the averaging process stable, we employ the non-linear translation averaging with the angle error that shares similar spirits with \cite{wilson2014robust}. Assuming $\mathbf{t}_{qd}^i$ is a unit vector, we have:

\begin{equation}
    \min \sum_i w_{qd}^id(\mathbf{R}_d^\mathrm{T}\mathbf{t}_{qd}^i, \frac{\mathbf{C}_q - \mathbf{C}_d}{|\mathbf{C}_q - \mathbf{C}_d|}),
\end{equation}

where $w_{qd}^i$ is the weight of the image pair depending on the number of inlier matches. In practice, we optimize this non-linear problem on the inlier image pairs with less than 5 degree of consistency with the output rotation after rotation averaging. Note that the non-linear refinement works better for us because it is not affected by the distance between the two queries, while other methods based on 3D point-line distances \cite{ozyesil2015robust} have a bias since zero error is achieved when the two camera centers collapse into the same point, which makes it particular unstable for multi-query co-localization. 

\subsection{Post Optimization} \label{sec:ba}
After rotation averaging and translation averaging, we get a rough estimation of the global rotation and translation of the query image. To further improve its accuracy, we perform a post optimization on all the inlier image pairs. We collect the corresponding inlier matches in those pairs, and conduct the optimization over all the point matches. 

A naive solution is to optimize the Sampson error \cite{hartley2003multiple} for each pair of correspondences. While this already employs richer information than the input relative pose itself, it cannot handle the degenerate case with collinear motions on the database images. Actually, this degenerate case cannot be handled if we only use the two-view constraints, because all positions along the line will result in zero error. However, in our case, since the poses from the database are available, we can use it to triangulate 3D point to fix the scale on the degenerate direction. While simply performing exhaustive feature matching and point triangulation on the fly as in \cite{sattler2017large} is inefficient, we can make use of the inlier matches between the query image and multiple database images to form a feature track over each keypoint of the query image. In this way, we can get feature track with more than one database images, which makes it possible to fix the scale problem under degenerate cases.

Specifically, for each keypoint of the query image we connect it to all the correspondingly matched keypoints in the database images that are considered as inliers in the two-view geometry estimation. Then, we triangulate one 3D point per track with multi-view triangulation \cite{schonberger2016structure} and perform joint optimization with fixed database over the points and the query pose to get the final output.

\subsection{Extensibility}

Since all the discussions on the three steps motion averaging are naturally generalizable, the described LazyLoc pipeline can be directly extended to more complex pose graph configurations by adding edge among queries, which enables localizing multiple queries together. This can integrate the advantages of motion tracking and rigs (illustrated in Figure~\ref{figure:extension}) to improve localization accuracy.

\input{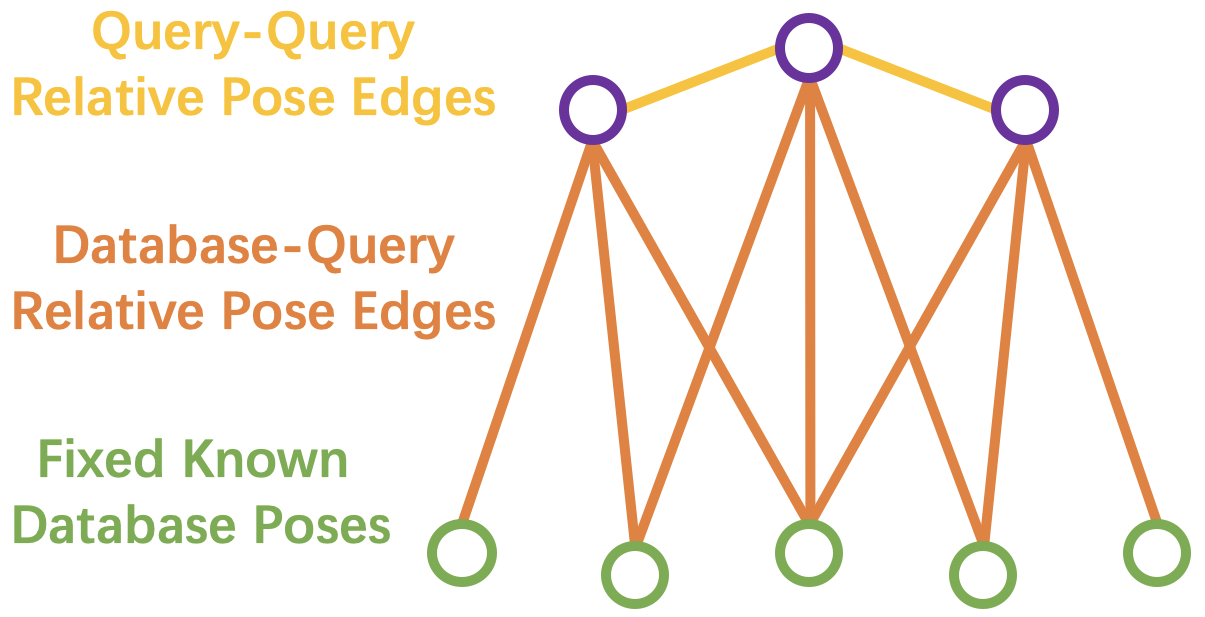}

%% file: fig/extension.tex
\begin{figure}[tb]
\centering
	\includegraphics[width=0.99\linewidth]{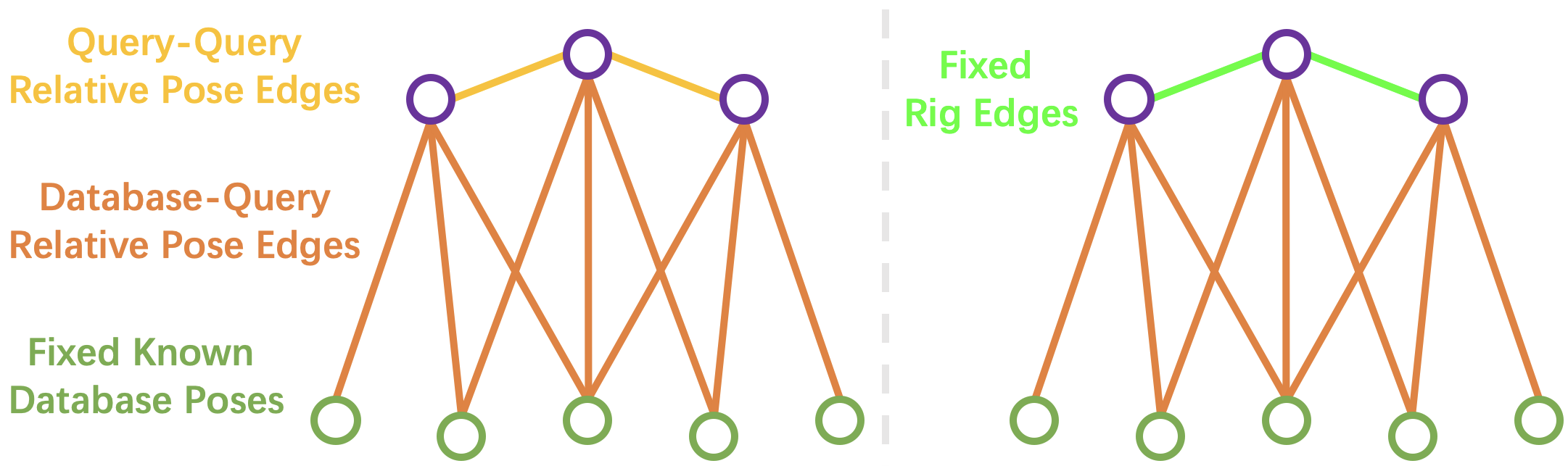}
	\caption{
    Illustration of extending LazyLoc to multi-query co-localization (left) and camera rigs (right). By simply adding edges among query images, the LazyLoc framework naturally extends to localize multiple images together. It is worth noting that the framework can also be extended to generalized camera models by simply fixing the relative poses among queries.
	}
	\label{figure:extension}
\end{figure}

%% file: tex/4_experiment.tex
\section{Experiments}
Below, we report the localization results of LazyLoc on several datasets.
The results consist of comparisons with state-of-the-art methods (Section~\ref{sec:main}), analyses (Section~\ref{sec:analyses}), and verifications of extensions (Section~\ref{sec:extensions}). 

\subsection{Main Experiments} \label{sec:main}

\noindent
\textbf{Datasets.} 
We conduct our main experiments on the 7 Scenes dataset~\cite{shotton2013scene}, Cambridge Landmarks~\cite{kendall2015posenet}, and Aachen Day-Night benchmark~\cite{Sattler2012BMVC, Sattler2018CVPR}. The 7 Scenes dataset consists of 7 small indoor scenes, the Cambridge dataset contains 5 outdoor scenes, while the Aachen dataset is a large-scale outdoor scene. 

~\\
\noindent
\textbf{Competitors.} 
For the accurate localization purpose, in the main experiments, we compare with only structure-based methods. 
Specifically, we select Active Search (AS)~\cite{sattler2016efficient}, InLoc~\cite{taira2018inloc}, HLoc~\cite{sarlin2019coarse, sarlin2020superglue}, and PixLoc~\cite{sarlin2021back} as representatives for SfM-based feature matching approaches, and select HSCNet~\cite{li2020hierarchical}, DSAC*~\cite{brachmann2021visual}, VS-Net~\cite{huang2021vs}, and DSM~\cite{tang2021learning} to represent the scene coordinate regression series. 
Active Search performs direct 2D-3D correspondence search, while InLoc, HLoc, and PixLoc apply image retrieval before local feature matching. 
HSCNet, DSAC*, and VS-Net require training in each scene, while DSM is scene agnostic. 
Note that all of the aforementioned methods require pre-built 3D metric maps either explicitly, or implicitly as 3D coordinates or depths. 
In addition, we compare with a baseline~\cite{zhou2020learn}, which represents the state-of-the-art solution without pre-built 3D metric maps.

~\\
\noindent
\textbf{Implementation details.}
Following HLoc, We apply NetVLAD~\cite{arandjelovic2016netvlad}, SuperPoint~\cite{detone2018superpoint}, and SuperGlue~\cite{sarlin2020superglue} for image retrieval, keypoint extraction, and matching, respectively.
For the main experiments, we by default use top-20 retrieved images for 7 Scenes and Cambridge, and top-50 images for Aachen, which are similar to HLoc.

\input{tab/7s}

\input{tab/cambridge}

\input{tab/7s_sift}
\input{tab/cambridge_sift}

~\\
\noindent
\textbf{7 Scenes main results.} 
The results are shown in Table~\ref{tab:7s}.
We first compare with SfM-based feature matching approaches and our method achieves lower pose errors. It is mainly because the 3D metric maps built from databases contain inevitable errors during the 3D point triangulation process, even using ground truth camera poses. 
Regarding scene coordinate regression, our method slightly lags behind HSCNet, DSAC*, and VS-Net. This is because these competitors take time to train individual network parameters for each specific scene, which is a heavy and inefficient process. Although DSM is designed scene agnostic, it requires a depth image for each database image to infer 3D coordinates, while we require only RGB images. Even though, our pose errors are lower than DSM. 
Another noticeable case is the scene Stairs that contains large areas of repetitive and textureless regions, which are quite challenging for feature matching based approaches. As a result, the SfM-based methods are overall worse than scene coordinate regression. We are glad to see that we obtain a reasonable result in this scene.
To conclude, the results on the 7 Scenes dataset demonstrate that using the proposed motion averaging framework, one can achieve comparable localization accuracy with state-of-the-art methods, without resorting to pre-built 3D metric maps.

~\\
\noindent
\textbf{Cambridge main results.} 
The database images on Cambridge are sparser than those on 7 Scenes. Also, there are collinear motions that are in general challenging for common motion averaging methods. 
However, we still reach comparable accuracy with the state-of-the-art. 
As shown in Table~\ref{tab:cambridge}, our average pose errors rank the second-best among all the competitors, only behind HLoc*. 
This surprising finding shows that visual localization without pre-built 3D metric maps can achieve very competitive performance compared to structure-based methods. 
It also confirms the effectiveness and robustness of the proposed localization system, thanks to the carefully designed robust motion averaging. 
The detailed study and analysis of our motion averaging framework is conducted in Section~\ref{sec:analyses}.
While scene coordinate regression methods achieve promising results on 7 scenes, they struggle on Cambridge in most of the scenes. This is probably due to the fact that the dataset does not provide ground truth depth images.
Therefore, Multi-View Stereo (MVS) algorithms are applied to recover dense 3D maps, as described in recent work~\cite{brachmann2018learning}. As a result, the pose estimation is affected by errors in the 3D reconstruction. 
In contrast, our proposed method achieves reasonable accuracy, while it is convenient to deploy since it does not require the pre-processing to reconstruct 3D metric maps from database images.

\noindent
\textbf{Comparison with the baseline~\cite{zhou2020learn}.} 
Besides structure-based methods, we compare our method with the state-of-the-art localization method that does not require pre-built 3D metric maps. 
For a fair comparison, we apply a similar experimental setting as in \cite{zhou2020learn} that uses only top-5 retrieved database images, SIFT keypoints \cite{lowe1999object}, and a naive nearest neighbor matcher, denoted by LazyLoc*. 
As a result, there are fewer relative poses available for absolute pose optimization, and their accuracy becomes much worse. The results of the final absolute pose errors are shown in Table \ref{tab:7s_sift} and \ref{tab:cambridge_sift}. As expected, we observe that our pose errors are significantly larger than those in Table~\ref{tab:7s} and \ref{tab:cambridge}. Even though, LazyLoc* still outperforms the baseline. 
It demonstrates that applying robust motion averaging is better than the conventional RANSAC scheme on top of relative poses. The detailed analysis of motion averaging that replaces our post optimization with local optimization is reported in Section~\ref{sec:analyses}. 
Not only is our method better than theirs on the motion averaging part, but we also propose a post optimization scheme that further improves the performance.

~\\
\noindent
\textbf{Aachen Day-Night.} 
Scene coordinate regression approaches can hardly handle large-scale outdoor scenes. ESAC~\cite{brachmann2019expert} and HSCNet are the only two from scene coordinate regression series that are submitted to the Aachen benchmark. As shown in Table~\ref{tab:aachen}, they lag behind the SfM-based methods, especially in the Night group. 
This is because the scene coordinate regression methods essentially memorize the training data (in the day time) so that they can hardly handle un-trained query conditions (in the night time). 
We are glad to see that we outperform the original Active Search and PixLoc, achieving very close accuracy to HLoc. 
Note that the Aachen dataset contains repetitive patterns and similar buildings. Therefore, the image retrieval process may return wrong database images, resulting in failure cases. When there is only one valid database image available, our method can not solve the scale factor from the single relative pose, while PnP-based methods using 3D maps can solve the absolute pose. In addition, optimizing an accurate pose from only a few relative poses is still very challenging for motion averaging. This points out the main limitation of our proposed method - the minimal configuration requires 2 retrieved database images, and the fewer the images the lower robustness. 
\input{tab/aachen}

~\\
\subsection{Analyses} \label{sec:analyses} 

In this section, we present a series of analyses of our proposed visual localization system. 
First, we perform ablation studies to evaluate the effectiveness of the motion averaging applied in our system. 
Then, we conduct experiments to demonstrate how database errors and the top-$K$ images affect the system's performance. 
Finally, we discuss the computational complexity of our localization system. 

\input{tab/ablation_po}

~\\
\noindent
\textbf{Motion averaging.}
The results of our ablation studies are shown in Table~\ref{tab:ablation_po}. 
We first replace our translation averaging algorithm with LUD~\cite{ozyesil2015robust}. 
On the two datasets, we can observe that the results of LUD are consistently worse than our default non-linear optimization. 
It is because that LUD computes the point-line distances, which are sensitive to depth changes. And when two poses collapse, the error will be zero. 
Therefore, we leverage non-linear optimization on top of angle errors instead. 

We also show the results of removing our post optimization and replacing it with local optimization as described in \cite{zhou2020learn}, as well as using conventional Sampson error \cite{hartley2003multiple}. 
From the results, we observe that the proposed post optimization significantly contributes to our method. The poses become worse with the naive local optimization scheme. 
Note that the replacement with the Sampson error overall hurts the accuracy. It is mainly because of total failure in degenerate cases with colinear motions. Since the post optimization potentially integrates the track with $\geq$2 known database poses, we are robust to these challenging cases.

~\\
\noindent
\textbf{Noisy database.} 
To perform sensitivity analysis on errors of database poses, we add random noise to the retrieved poses. Within the noise range shown in Table~\ref{tab:noisy}, our system is robust and produces reasonable results. Noise in rotation has a large impact, since rotation estimation is the first step of our motion averaging module, and translation estimation and post optimization depend heavily on good rotation initialization. 
To showcase the impact of SfM triangulation
errors, we report the results of HLoc on the 7 Scenes dataset, using the SfM map and ground-truth map (built with depth images), respectively. The average median error decreases from 3.14cm / 1.09$^\circ$ to 2.71cm / 0.93$^\circ$. 

~\\
\noindent
\textbf{Top-$K$ images.} The localization results are more accurate if the retrievals are more correct. As shown in Table~\ref{tab:topk}, we observe that as the numbers of retrieved images increase, the proposed method obtains lower pose errors. 
The visual overlap, parallax, and extreme shifts among database and query images affect the relative pose estimation from two-view geometry. However, as long as there are enough reasonable relative poses, as shown in Figure~\ref{figure:vis}, our system successfully rejects outliers using robust motion averaging. 

\input{tab/noisy_db_pose}
\input{tab/topk}

\input{fig/vis}

~\\
\noindent
\textbf{Computational complexity.} 
Our feature extraction and matching modules are the same as those used by HLoc, the state-of-the-art structure-based competitor.
In addition, our pose solver (relative poses and motion averaging) takes about 400ms (Intel Xeon Platinum 8255C CPU @ 2.50GHz) for single queries.
Our post optimization module is similar to online local SfM \cite{sattler2017large, Sattler2018CVPR}, but it avoids exhaustive matching among database images ($\mathcal{O}(K^2)$). Instead, it only matches features on database-query pairs ($\mathcal{O}(K)$), which is more efficient. 


\vspace{10pt}

\subsection{Extensions} \label{sec:extensions}

~\\
\noindent
\textbf{Multi-query co-localization.}  
The test images from the 7 Scenes and Cambridge datasets are from videos. Therefore, we simply split the original image streams into fragments with different lengths, and solve the multiple camera poses inside each fragment simultaneously. The results are shown in Table~\ref{tab:multi}. We are glad to see that the multi-query co-localization is effective and slightly reduces pose errors. 

~\\
\noindent
\textbf{Camera rigs.}
We demonstrate an extension of generalized camera models by applying virtual camera rigs to the Cambridge dataset. Specifically, we regard $L$ consecutive images in each test sequence as a camera rig with known relative poses. The localization results are shown in Table~\ref{tab:cam_rigs}. 
Using fixed known query-query relative poses as constraints can consistently lead to lower pose errors.

\input{tab/multi}

\input{tab/cambridge_rigs}

%% file: tab/7s.tex
\begin{table*}[tb]
\centering
\resizebox{.99\textwidth}{!}{
\begin{tabular}{lcc|ccccccc|c}
\toprule
Error (cm / \textdegree) $\downarrow$ & Pre-Built 3D & Image & Chess & Fire & Heads & Office & Pumpkin & RedKitchen & Stairs & Average \\

& Metric Map & Type &  &  &  &  &  &  &  &  \\

\toprule

AS~\cite{sattler2016efficient} & SfM & RGB & 3 / 0.87 & 2 / 1.01 & 1 / 0.82 & 4 / 1.15 & 7 / 1.69 & 5 / 1.72 & 4 / 1.01 & 3.71 / 1.18 \\

InLoc~\cite{taira2018inloc} & RGB-D Fusion & RGB-D & 3 / 1.05 & 3 / 1.07 & 2 / 1.16 &  3 / 1.05 & 5 / 1.55 & 4 / 1.31 & 9 / 2.47 & 4.14 / 1.38 \\

HLoc~\cite{sarlin2019coarse, sarlin2020superglue} & SfM & RGB & 2 / 0.85 & 2 / 0.94 & 1 / 0.75 & 3 / 0.92 & 5 / 1.30 & 4 / 1.40 & 5 / 1.47 & 3.14 / 1.09 \\

PixLoc~\cite{sarlin2021back} & SfM & RGB & 2 / 0.80 & 2 / 0.73 & 1 / 0.82 & 3 / 0.82 & 4 / 1.21 & 3 / 1.20 & 5 / 1.30 & 2.86 / 0.98 \\

\midrule

HSCNet~\cite{li2020hierarchical} & Depth Fusion & RGB-D & 2 / 0.7 & 2 / 0.9 & 1 / 0.9 & 3 / 0.8 & 4 / 1.0 & 4 / 1.2 & 3 / 0.8 & 2.71 / 0.90 \\
DSAC*~\cite{brachmann2021visual} & SfM & RGB & 2 / 1.10 & 2 / 1.24 & 1 / 1.82 & 3 / 1.15 & 4 / 1.34 & 4 / 1.68 & 3 / 1.16 & 2.71 / 1.36 \\
VS-Net~\cite{huang2021vs} & Depth Fusion & RGB-D & 1.5 / 0.5 & 1.9 / 0.8 & 1.2 / 0.7 & 2.1 / 0.6 & 3.7 / 1.0 & 3.6 / 1.1 & 2.8 / 0.8 & 2.4 / 0.8 \\
DSM~\cite{tang2021learning} & Not Required & RGB-D & 2 / 0.71 & 2 / 0.85 & 1 / 0.85 & 3 / 0.84 & 4 / 1.16 & 4 / 1.17 & 5 / 1.33 & 3 / 0.99 \\

\midrule
LazyLoc (Ours) & Not Required & RGB & 2.45 / 0.81 & 2.19 / 0.94 & 1.04 / 0.66 & 2.14 / 0.65 & 4.20 / 0.88 & 3.01 / 1.24 & 4.79 / 1.48 & 2.83 / 0.95 \\
\bottomrule
\end{tabular}}
\caption{Main localization results on the 7 Scenes dataset. We report median translation and rotation errors. }
\label{tab:7s}
\end{table*}

%% file: tab/cambridge.tex
\begin{table*}[tb]
\centering
\resizebox{.96\textwidth}{!}{
\begin{tabular}{lcc|ccccc|c}
\toprule
Error (cm / \textdegree) $\downarrow$ & Pre-Built 3D & Image & GreatCourt & KingsCollege & OldHospital  & ShopFacade  & StMarysChurch & Average \\

& Metric Map & Type &  &  &  &  &  &  \\ 

\toprule

AS~\cite{sattler2016efficient} & SfM & RGB & 24 / 0.13    & 13 / 0.22    & 20 / 0.36    & 4 / 0.21    & 8 / 0.25      & 13.80 / 0.23         \\
PixLoc~\cite{sarlin2021back} & SfM & RGB & 30 / 0.14    & 14 / 0.24    & 16 / 0.32    & 5 / 0.23    & 10 / 0.34     & 15 / 0.25            \\
HLoc* & SfM & RGB & 9.6 / 0.05   & 6.5 / 0.10   & 12.6 / 0.23  & 2.9 / 0.14  & 4.4 / 0.16    & 7.20 / 0.14          \\

\midrule

HSCNet~\cite{li2020hierarchical} & MVS & RGB & 28 / 0.2     & 18 / 0.3     & 19 / 0.3     & 6 / 0.3     & 9 / 0.3       & 16 / 0.28            \\
DSAC*~\cite{brachmann2021visual} & SfM & RGB & 49 / 0.3     & 15 / 0.3     & 21 / 0.4     & 5 / 0.3     & 13 / 0.4      & 21 / 0.34            \\
VS-Net~\cite{huang2021vs} & MVS & RGB & 22 / 0.1     & 16 / 0.2     & 16 / 0.3     & 6 / 0.3     & 8 / 0.3       & 13.60 / 0.24         \\
DSM~\cite{tang2021learning} & MVS & RGB & 44 / 0.23    & 19 / 0.36    & 24 / 0.39    & 7 / 0.38    & 12 / 0.35 & 21.2 / 0.34 \\

\midrule

LazyLoc (Ours) & Not Required & RGB & 14.37 / 0.08 & 7.04 / 0.13 & 19.87 / 0.37 & 4.17 / 0.15 & 5.74 / 0.18 & 10.24 / 0.18 \\

\bottomrule
\end{tabular}}
\caption{Main localization results on the Cambridge dataset. 
HLoc* refers to the updated version of HLoc~\cite{sarlin2019coarse, sarlin2020superglue}. 
}
\label{tab:cambridge}
\end{table*}

%% file: tab/7s_sift.tex
\begin{table}[tb]
\centering
\setlength{\tabcolsep}{9pt}
\begin{tabular}{lcc}
\toprule
Error (cm / \textdegree) $\downarrow$ & Baseline~\cite{zhou2020learn} & LazyLoc*        \\
\toprule
Chess            & 4 / 1.48  & \textbf{2.64 / 0.87} \\
Fire             & 5 / 1.62  & \textbf{3.02 / 1.01} \\
Heads            & 4 / 1.80  & \textbf{1.28 / 0.85} \\
Office           & 6 / 1.59  & \textbf{2.59 / 0.79} \\
Pumpkin          & 8 / 1.86  & \textbf{4.89 / 1.26} \\
RedKitchen       & 7 / 1.86  & \textbf{3.58 / 1.27} \\
Stairs           & 22 / \textbf{3.69} & \textbf{21.17} / 4.63 \\
\midrule
Average          & 8 / 1.99  & \textbf{5.60 / 1.53} \\
\bottomrule
\end{tabular}
\caption{
Comparison with the baseline~\cite{zhou2020learn} under the same setting. The experiment is conducted on the 7 Scenes dataset. 
}
\label{tab:7s_sift}
\end{table}

%% file: tab/cambridge_sift.tex
\begin{table}[tb]
\centering
\setlength{\tabcolsep}{9pt}
\begin{tabular}{lcc}
\toprule
Error (cm / \textdegree) $\downarrow$ & Baseline~\cite{zhou2020learn} & LazyLoc*        \\
\toprule
GreatCourt    & -         & 63.05 / 0.25 \\
KingsCollege  & 48 / 0.71 & \textbf{13.22 / 0.23} \\
OldHospital   & 88 / \textbf{1.24} & \textbf{79.24} / 1.28 \\
ShopFacade    & 17 / 0.55 & \textbf{6.26 / 0.30}  \\
StMarysChurch & 35 / 1.00 & \textbf{19.32 / 0.65} \\
\midrule
Average       & -         & 36.22 / 0.54 \\
Average*      & 47 / 0.88 & \textbf{29.51 / 0.62} \\
\bottomrule
\end{tabular}
\caption{
Comparison with the baseline~\cite{zhou2020learn} under the same setting. The experiment is conducted on the Cambridge dataset. 
Average* denotes the average of the 4 scenes except for GreatCourt.
}
\label{tab:cambridge_sift}
\end{table}

%% file: tab/aachen.tex
\begin{table}[tb]
\centering
\resizebox{.48\textwidth}{!}{
\begin{tabular}{lc|cc}
\toprule
Accuracy (\%) $\uparrow$ & Pre-Built 3D & Day & Night \\
 & Metric Map &  &  \\
\toprule

AS~\cite{sattler2016efficient} & SfM & 57.3 / 83.7 / 96.6 & 28.6 / 37.8 / 51.0 \\
AS v1.1 & SfM & 85.3 / 92.2 / 97.9 & 39.8 / 49.0 / 64.3 \\
HLoc~\cite{sarlin2019coarse, sarlin2020superglue} & SfM & 89.6 / 95.4 / 98.8 & 86.7 / 93.9 / 100  \\
PixLoc~\cite{sarlin2021back} & SfM & 68.0 / 74.6 / 80.8 & 57.1 / 69.4 / 76.5 \\

\midrule

ESAC~\cite{brachmann2019expert} & SfM & 42.6 / 59.6 / 75.5 & 6.1 / 10.2 / 18.4  \\
HSCNet~\cite{li2020hierarchical} & SfM & 71.1 / 81.9 / 91.7 & 32.7 / 43.9 / 65.3 \\

\midrule

LazyLoc (Ours) & Not Required & 85.4 / 92.7 / 95.8 & 80.6 / 88.8 / 94.9 \\
\bottomrule
\end{tabular}}
\caption{Localization results on the Aachen dataset. }
\label{tab:aachen}
\end{table}

%% file: tab/ablation_po.tex
\begin{table}[tb]
\centering

\setlength{\tabcolsep}{6pt}
\begin{tabular}{lcc}
\toprule
Error (cm / \textdegree) $\downarrow$ & 7 Scenes & Cambridge \\
\toprule
LazyLoc w/ LUD~\cite{ozyesil2015robust} & 3.35 / 1.06 & 21.94 / 0.34 \\

\midrule

LazyLoc w/o Opt. & 4.10 / 1.23 & 35.74 / 0.58 \\
LazyLoc w/ LO~ in \cite{zhou2020learn} & 5.58 / 1.24 & 55.87 / 0.67 \\
LazyLoc w/ Sampson~\cite{hartley2003multiple} & 4.36 / 1.39 & 41.42 / 0.78 \\

\midrule

LazyLoc (Full) & \textbf{2.83 / 0.95} & \textbf{10.24 / 0.18} \\
\bottomrule
\end{tabular}

\caption{Ablation studies of motion averaging. }
\label{tab:ablation_po}
\end{table}




%% file: tab/noisy_db_pose.tex
\begin{table}[tb]
\centering
\setlength{\tabcolsep}{9pt}

\resizebox{.48\textwidth}{!}{

\begin{tabular}{l|c|c|c|c}
\hline

Rot. noise ($^\circ$) & 0 & 1 & 5 & 10 \\ 
\hline
Error (cm / $^\circ$) $\downarrow$ & 2.83 / 0.95 & 3.64 / 1.48 & 11.92 / 5.14 & 23.17 / 9.98 \\ 

\hline

Transl. noise (cm) & 0 & 1 & 5 & 10 \\ 
\hline
Error (cm / $^\circ$) $\downarrow$ & 2.83 / 0.95 & 3.19 / 0.97 & 6.24 / 1.14 & 11.01 / 1.35 \\ 

\hline
\end{tabular}

}

\caption{
Localization results based on noisy database poses. 
The experiment is conducted on the 7 Scenes dataset.
}
\label{tab:noisy}
\end{table}


%% file: tab/topk.tex
\begin{table}[tb]
\centering
\resizebox{.48\textwidth}{!}{
\begin{tabular}{l|c|c|c|c}
\hline
Top-K retrieval & 2 & 5 & 10 & 20 \\ 
\hline
Error (cm / $^\circ$) & 5.64 / 1.42 & 3.36 / 1.05 & 2.91 / 0.97 & 2.83 / 0.95 \\ 
\hline
\end{tabular}
}
\caption{
Localization results using different numbers of retrieved images. 
The experiment is conducted on the 7 Scenes dataset.
}
\label{tab:topk}
\end{table}

%% file: fig/vis.tex
\begin{figure}[!tbp]
\centering
\includegraphics[width=1.0\linewidth]{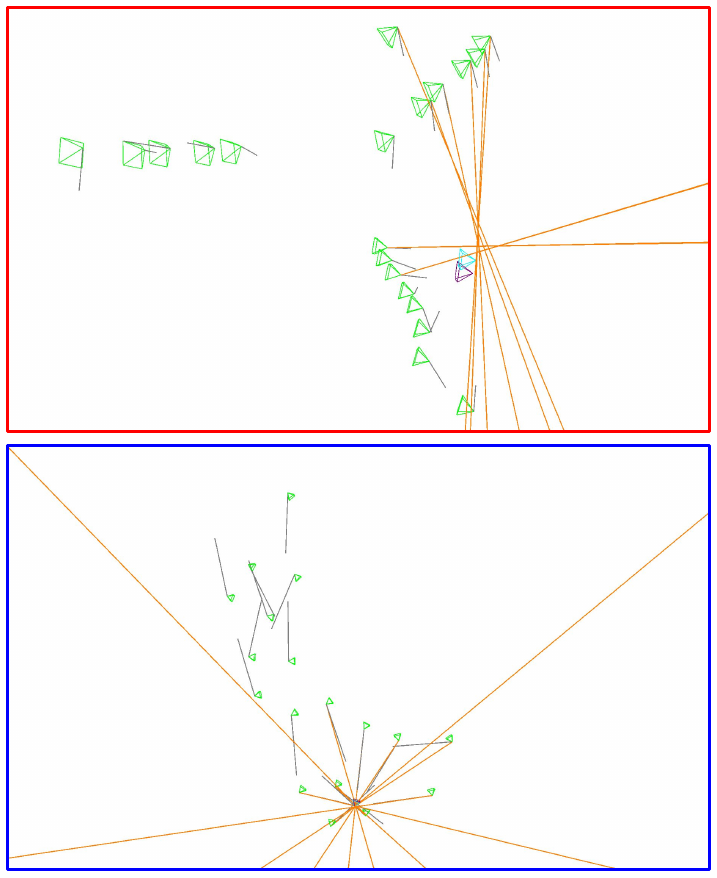}
\caption{ Visualization of our localization results. We randomly pick two cases lying around the median pose error from the 7 Scenes (top) and Cambridge (bottom) datasets, respectively. We visualize the retrieved database poses (green frustums), the relative translation vectors (gray segments), the optimized inlier translation vectors after motion averaging (orange rays), the final pose estimation (purple frustum), and the ground truth (cyan frustum). }
\label{figure:vis}
\end{figure}

%% file: tab/multi.tex
\begin{table}[h]
\centering
\setlength{\tabcolsep}{8pt}

\resizebox{.48\textwidth}{!}{

\begin{tabular}{lccc}
\toprule
Error (cm / \textdegree) $\downarrow$ & LazyLoc (Single) & LazyLoc (Length=3) & LazyLoc (Length=5) \\
\toprule
7 Scenes & 2.83 / 0.95 & 2.78 / 0.95 & \textbf{2.75} / \textbf{0.94} \\

\hline

Cambridge & 10.24 / 0.18 & 9.90 / 0.19 & \textbf{9.47} / \textbf{0.17} \\ 
\bottomrule
\end{tabular}

}

\caption{Results of multi-query co-localization. }
\label{tab:multi}
\end{table}


%% file: tab/cambridge_rigs.tex
\begin{table}[h]
\centering
\resizebox{.48\textwidth}{!}{
\begin{tabular}{lcccc}
\toprule
Error (cm / °) $\downarrow$ & MQ ($L=3$) & CR ($L=3$) & MQ ($L=5$) & CR ($L=5$) \\
\toprule
GreatCourt & 11.32 / 0.06 & 8.68 / 0.04 & 10.44 / 0.05 & 7.08 / 0.03 \\
KingsCollege & 9.16 / 0.17 & 4.15 / 0.08 & 8.92 / 0.18 & 3.47 / 0.07 \\
OldHospital & 20.42 / 0.42 & 9.22 / 0.16 & 19.95 / 0.36 & 7.53 / 0.19 \\
ShopFacade & 3.15 / 0.13 & 1.67 / 0.07 & 3.61 / 0.13 & 1.40 / 0.05 \\
StMarysChurch & 5.46 / 0.16 & 2.92 / 0.09 & 4.45 / 0.15 & 2.28 / 0.08  \\
\midrule
Average & 9.90 / 0.19 & 5.33 / 0.09 & 9.47 / 0.17 & 4.35 / 0.08 \\
\bottomrule
\end{tabular}
}
\caption{Demonstration of camera rigs on the Cambridge dataset. MQ and CR denote multi-query and camera rigs, respectively. }
\label{tab:cam_rigs}
\end{table}

%% file: tex/5_conclusion.tex
\section{Conclusion}

In this technical report, we present surprising findings that a visual localization system without pre-built 3D metric maps can be nearly as competitive as state-of-the-art structure-based methods. We believe that this opens up new perspectives to the visual localization community. Future research directions for the proposed method include extending it with advanced two-view geometry approaches, as well as incorporating diverse geometry structures such as line segments and vanishing points. It could also be improved to handle dynamic scenes with changing conditions.